\documentclass[10pt,twocolumn,letterpaper]{article}

\usepackage[pagenumbers]{cvpr} 

\definecolor{cvprblue}{rgb}{0.21,0.49,0.74}
\usepackage[pagebackref,breaklinks,colorlinks,allcolors=cvprblue]{hyperref}

\usepackage{pifont}
\usepackage{multirow}
\usepackage[normalem]{ulem}
\newcommand{\cmark}{\ding{51}}%
\newcommand{\xmark}{\ding{55}}%
\usepackage{amssymb}
\usepackage{amsmath}
\usepackage{booktabs}       
\usepackage{amsfonts}       
\usepackage{nicefrac}       
\usepackage{xcolor}         
\usepackage[table]{xcolor}
\usepackage{bbm}

\usepackage{microtype}      

\def\paperID{12295} 
\def\confName{CVPR}
\def\confYear{2026}

\title{SurgMLLMBench: A Multimodal Large Language Model Benchmark Dataset \\for Surgical Scene Understanding}

\author{
Tae-Min Choi\textsuperscript{1} \quad
Tae Kyeong Jeong\textsuperscript{2,*} \quad 
Garam Kim\textsuperscript{2,*} \quad 
Jaemin Lee\textsuperscript{3} \quad
Yeongyoon Koh\textsuperscript{4} \\
In Cheul Choi\textsuperscript{4} \quad
Jae-Ho Chung\textsuperscript{3} \quad
Jong Woong Park\textsuperscript{4} \quad
Juyoun Park\textsuperscript{2,†}\\
\textsuperscript{1}Samsung Research \quad
\textsuperscript{2}Center for Humanoid Research, Korea Institute of Science and Technology \\
\textsuperscript{3}Department of plastic surgery, College of medicine, Korea University \\
\textsuperscript{4}Department of orthopedic surgery, College of medicine, Korea University \\
}

\begin{document}
\twocolumn[{%
\renewcommand\twocolumn[1][]{#1}%
\maketitle
    \vspace{-1.7em}
    \centering
    \includegraphics[width=0.95\linewidth, height=7cm]{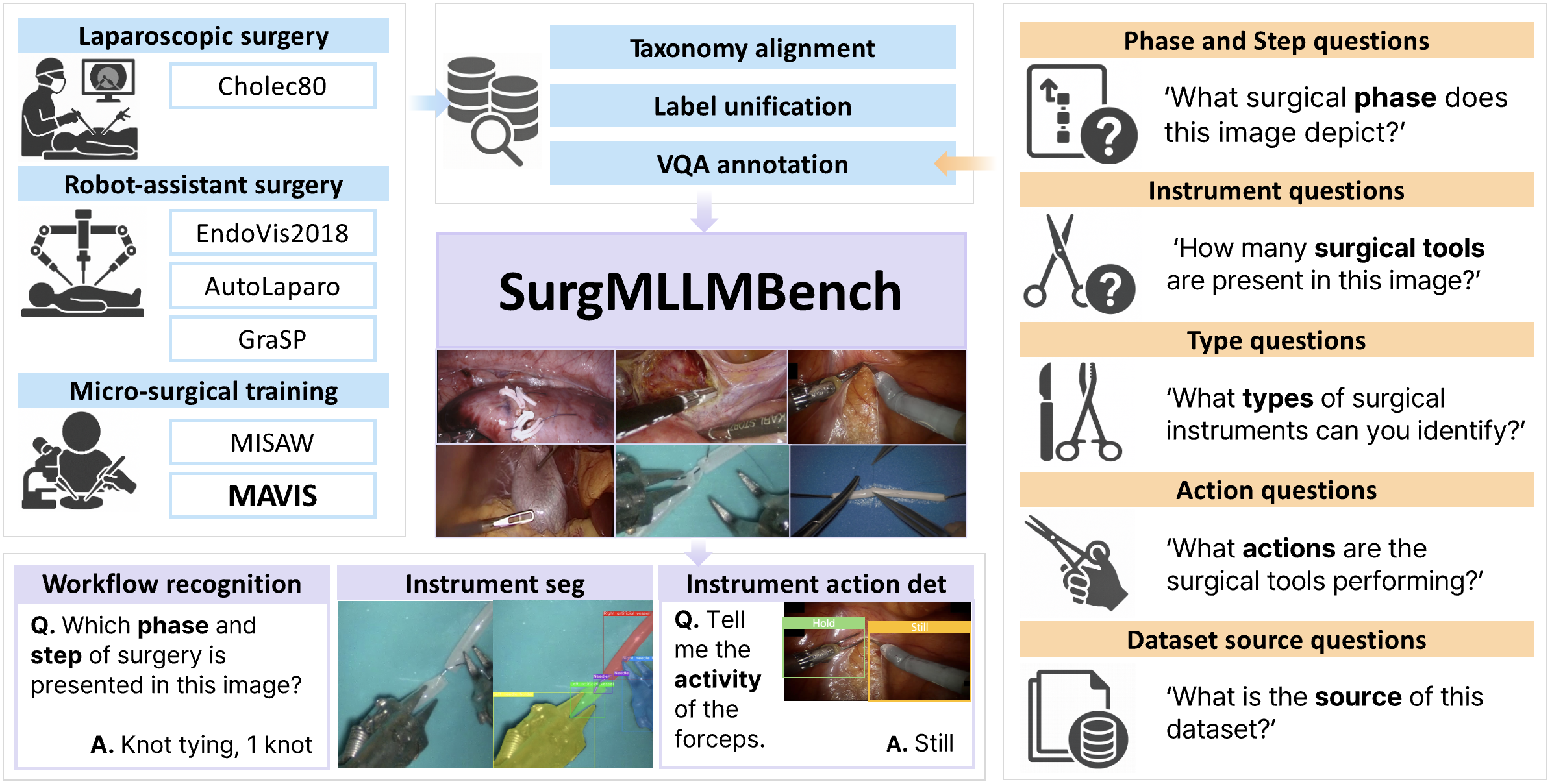}
    \captionof{figure}{Overview of \textbf{SurgMLLMBench}. Multi-domain surgical datasets, including the newly collected \textbf{MAVIS}, are unified into a multimodal benchmark through taxonomy alignment, label unification, and VQA annotation. The template-based question generator (orange) produces structured VQA pairs using five query types. SurgMLLMBench supports interactive multimodal surgical scene understanding.
    \vspace{2em}}
    \label{fig:surgmllmbench_overview}
}]


\begin{abstract}
Recent advances in multimodal large language models (LLMs) have highlighted their potential for medical and surgical applications. However, existing surgical datasets predominantly adopt a Visual Question Answering (VQA) format with heterogeneous taxonomies and lack support for pixel-level segmentation, limiting consistent evaluation and applicability. We present SurgMLLMBench, a unified multimodal benchmark explicitly designed for developing and evaluating interactive multimodal LLMs for surgical scene understanding, including the newly collected Micro-surgical Artificial Vascular anastomosIS (MAVIS) dataset. It integrates pixel-level instrument segmentation masks and structured VQA annotations across laparoscopic, robot-assisted, and micro-surgical domains under a unified taxonomy, enabling comprehensive evaluation beyond traditional VQA tasks and richer visual–conversational interactions. Extensive baseline experiments show that a single model trained on SurgMLLMBench achieves consistent performance across domains and generalizes effectively to unseen datasets. SurgMLLMBench will be publicly released as a robust resource to advance multimodal surgical AI research, supporting reproducible evaluation and development of interactive surgical reasoning models.\footnote{http://surgmllmbench.github.io/}
\end{abstract}

\section{Introduction}
In minimally invasive procedures such as laparoscopic and robot-assisted surgery, surgeons depend almost entirely on the restricted, two-dimensional view provided by an endoscopic camera. Because the camera can rotate, zoom, or be obscured by instruments, maintaining real-time awareness of the procedural stage and of each tool’s precise action is challenging even for experienced surgeons \cite{mascagni2022computer, ramesh2023dissecting}. Multimodal large language models (LLMs) \cite{liu2023visual} offer a promising route to surgical assistance: by jointly interpreting images and text, a model could describe the ongoing step, highlight relevant instruments on the display, and answer intra-operative queries as they arise. Achieving this capability, however, requires training data that links rich workflow context with fine-grained visual localization \cite{xu2025surgical, li2025surgical, zhu2024}.

Existing publicly available datasets have been developed primarily for Visual Question Answering (VQA) \cite{seenivasan2022surgical, yuan2024advancing, bai2025surgical} tasks and therefore do not capture the full range of information needed for comprehensive scene understanding. EndoChat \cite{wang2025endochat} and LLaVA-Surg \cite{li2024llava} provide hundreds of thousands of question–answer pairs, yet omit annotations for the surgical phase and procedural steps, leaving the temporal structure of an operation unrepresented. Surgical-LLaVA \cite{jin2024surgical} supplies phase labels, but its spatial supervision is limited to bounding-boxes rather than pixel-level delineations. The absence of detailed workflow annotations and fine-grained localization restricts how multimodal LLMs can be trained or evaluated as effective intra-operative assistants.

To address these limitations, we present \textbf{SurgMLLMBench}, a benchmark expressly designed around the full spectrum of surgical scene-understanding tasks, from global workflow recognition to fine-grained visual grounding. SurgMLLMBench integrates complementary annotations for the surgical phase, procedural step, instrument-centered actions, and pixel-level instrument segmentation, thereby covering an operation's temporal and spatial aspects. The dataset further embraces domain diversity by combining laparoscopic surgery, robot-assisted surgery, and micro-surgical training procedures with sequences captured in simulated environments, improving domain shift robustness. Also, unlike prior VQA-centric corpora that offer only bounding-box supervision, SurgMLLMBench provides dense segmentation masks, enabling multimodal LLMs to ground their textual outputs in pixel-accurate visual evidence.
An overview of the proposed SurgMLLMBench is shown in \cref{fig:surgmllmbench_overview}.

The SurgMLLMBench benchmark unifies six surgical datasets—Cholec80 \cite{twinanda2016endonet}, EndoVis2018 \cite{allan20202018}, AutoLaparo \cite{wang2022autolaparo}, MISAW \cite{huaulme2021micro}, GraSP \cite{ayobi2024pixel}, and a newly collected \textbf{Micro-surgical Artificial Vascular anastomosIS (MAVIS)} dataset. MAVIS captures complete surgical sequences in micro-surgical training sessions and introduces a hierarchical workflow taxonomy with a novel attribute, providing detailed representations of surgical workflows. Each frame across all datasets is organized around four complementary tasks: (1) phase recognition, assigning coarse workflow phases to capture global context; (2) step classification, refining this context by identifying the specific procedural step underway; (3) instrument-centered action detection, labeling the functional primitive executed by each active tool (e.g., grasp, cut, suture) for fine-grained temporal reasoning; and (4) instrument segmentation, providing pixel-accurate masks for every visible instrument instance and offering the spatial detail required for precise visual grounding. By aligning all source videos to this standard set of tasks, SurgMLLMBench supplies a single supervision signal that spans the temporal structure and spatial layout of surgical scenes.

Because most current multimodal LLMs are limited to textual responses and coarse bounding-box outputs, we select OMG-LLaVA \cite{zhang2024omg} as our baseline model. OMG-LLaVA combines a general-purpose segmentation encoder with a vision–language decoder, enabling the system to generate pixel-level masks while simultaneously producing natural-language answers to surgical queries. When trained on SurgMLLMBench, the model exhibits stable performance across datasets from multiple domains and demonstrates generalization ability to adapt even to datasets that were not included in the training. 
These findings indicate that the dense masks supplied by SurgMLLMBench not only encourage multimodal LLMs to link textual explanations with pixel-accurate visual evidence but also contribute to their robust understanding of surgical workflows across diverse scenarios.
This paper contributes the SurgMLLMBench dataset, a reproducible integration pipeline that maps heterogeneous surgical resources into a coherent annotation scheme, and quantitative baselines that highlight both the benefits and the remaining challenges of applying interactive multimodal LLMs in surgical settings. 

\section{Related Work}
\subsection{Benchmarks for Surgical Scene Understanding}
Recent advancements in surgical AI have accelerated the development of various benchmark datasets aimed at evaluating surgical scene understanding models comprehensively \cite{khan2025}. The GraSP \cite{ayobi2024pixel} dataset addresses surgical scene understanding by providing hierarchical tasks including surgical phase recognition, procedural steps identification, and fine-grained tasks such as surgical instrument segmentation and atomic visual action detection specifically within robot-assisted prostatectomy scenarios. Additionally, the recently proposed MM-OR dataset \cite{ozsoy2025mm} captures multimodal surgical environments by incorporating RGB-D video, audio data, speech transcripts, robot log data, and semantic scene graphs, facilitating tasks such as panoptic segmentation and holistic operating room (OR) scene understanding.
Although these datasets significantly contribute to the surgical AI community by providing diverse annotations and tasks, they predominantly focus on specific surgical procedures or modalities, limiting their generalizability and application in broader surgical contexts. Moreover, many current benchmarks are confined to static question-answering frameworks, restricting their effectiveness in evaluating dynamic and interactive multimodal models for real-world surgical environments.

\begin{table*}[t]
\centering
\small
\renewcommand{\arraystretch}{0.8}
\begin{tabular}{lcccccccc}
\toprule
Dataset      & \begin{tabular}[c]{@{}c@{}}Stage \\ Recog.\end{tabular} & \begin{tabular}[c]{@{}c@{}}Phase \\ Recog.\end{tabular} & \begin{tabular}[c]{@{}c@{}}Step \\ Recog.\end{tabular} & \begin{tabular}[c]{@{}c@{}}Instrument \\ Action Recog.\end{tabular} & \begin{tabular}[c]{@{}c@{}}Instrument \\ Segmentation\end{tabular} & Video Hour & \begin{tabular}[c]{@{}c@{}}Total \\ Frames\end{tabular} & \begin{tabular}[c]{@{}c@{}}Data \\ Domain\end{tabular} \\ 
\midrule
Cholec80 \cite{twinanda2016endonet}     &  & \cmark &  &  &  & 51.25 & 184,498 & LS \\
EndoVis2018 \cite{allan20202018} &  &  &  & \cmark & \cmark & 1.58 & 2,235 & RAS \\
AutoLaparo \cite{wang2022autolaparo}  &  & \cmark &  &  & \cmark & 23.13 & 83,243 & RAS \\
GraSP \cite{ayobi2024pixel}       &  & \cmark & \cmark & \cmark & \cmark & 32.37 & 116,515 & RAS \\
MISAW\textsuperscript{\textdaggerdbl} \cite{huaulme2021micro}        &  & \cmark & \cmark & \cmark & \cmark & 1.52 & 164,275 & MST \\
\textbf{MAVIS}       & \cmark & \cmark & \cmark &  & \cmark & 2.95 & 10,652 & MST \\ 
\midrule
Total Annotations & 10,652 & 559,182 & 291,442 & 31,872 & 25,221 & 112.80 & 561,418 &  \\ 
\bottomrule
\end{tabular}
\caption{Comparison of datasets. A checkmark (\cmark) indicates the presence of corresponding annotations. 
(Recog. denotes recognition. MISAW\textsuperscript{\textdaggerdbl} comprises the original MISAW dataset and supplementary segmentation annotations from \cite{Jeong2025MicrosurgicalIS}.)}
\label{tab:dataset_comparison}
\end{table*}

\subsection{Multimodal LLMs in Surgical Applications}
Multimodal LLMs have recently demonstrated significant promise across various medical and surgical applications by integrating visual, textual, and contextual data. EndoChat \cite{wang2025endochat} utilizes the large-scale multimodal Surg-396K dataset to enable interactive surgical education, supporting complex interactions through combined visual and textual queries. LLaVA-Surg \cite{li2024llava} introduces the Surg-QA dataset, generated from surgical lecture videos, to train a conversational vision-language assistant capable of answering open-ended queries regarding surgical scenarios. Similarly, Surgical-LLaVA \cite{jin2024surgical} enhances multimodal interactions by integrating language models with visual encoders specifically trained for surgical spatiotemporal contexts, demonstrating improved performance in surgical video understanding tasks.
Despite these advances, existing multimodal surgical LLMs largely remain restricted to predefined static interaction paradigms, lacking the ability to adapt dynamically to evolving surgical conditions. Particularly, tasks involving detailed pixel-level segmentation and real-time recognition of procedural contexts have been inadequately addressed due to the absence of richly annotated interactive datasets.


\section{SurgMLLMBench Dataset}
In this section, we introduce the SurgMLLMBench, a new benchmark recognizing the overall surgical procedure, scene information, and tool segmentation abilities for multimodal LLMs. We utilize five existing open-source surgical datasets and a newly generated dataset.
An overview of the overall dataset composition and benchmark structure is illustrated in \cref{fig:surgmllmbench_overview}.
To ensure compatibility across data sources, all six datasets were standardized under a unified annotation schema, and supplementary prompt-level annotations were introduced to support multimodal LLM training. 
The SurgMLLMBench dataset is publicly available at: \href{https://huggingface.co/datasets/KIST-HARILAB/SurgMLLMBench}{huggingface.co/datasets/KIST-HARILAB/SurgMLLMBench}.

\subsection{Data Collection}
\noindent\textbf{Public Datasets.} We construct an integrated benchmark dataset for training multimodal LLMs including five widely used surgical video datasets: Cholec80 \cite{twinanda2016endonet}, EndoVis2018 \cite{allan20202018}, AutoLaparo \cite{wang2022autolaparo}, GraSP \cite{ayobi2024pixel}, and MISAW \cite{huaulme2021micro}. 
Since the original MISAW dataset does not include instrument segmentation annotations, we incorporate supplementary segmentation data containing approximately 3,000 annotated images provided by \cite{Jeong2025MicrosurgicalIS}. 
To leverage the cross-domain characteristics in SurgMLLMBench, we included surgical video datasets covering diverse procedures and clinical domains. We categorize the datasets according to the surgical environment, distinguishing them as Laparoscopic Surgery (LS), Robot-Assisted Surgery (RAS), or Micro-Surgical Training procedures (MST) following \cite{ayobi2024pixel}. A detailed summary of this categorization is provided in \cref{tab:dataset_comparison}.

Since the existing datasets differ in resolution, frame rate, annotation format, and task definitions, a standardization process was required to integrate all videos and annotations into a unified structure. First, all videos were converted into frame-level representations, and a COCO-style \cite{lin2014microsoft} metadata schema was designed to harmonize annotation information across datasets. Each frame includes the fields: video ID, frame ID, stage, phase, step, instrument action, and segmentation. Missing entries were left blank to maintain structural consistency. This integration process was carefully designed to harmonize annotation standards across different sources while maintaining spatiotemporal consistency and semantic interpretability. 
Based on this unified structure, all datasets were reorganized to enable training and evaluation under a single, consistent format. The range of annotations, video durations, and surgical domains provided by each dataset are summarized in \cref{tab:dataset_comparison}, which illustrates the diversity and complexity of the tasks encompassed within SurgMLLMBench. The table visualizes the annotation coverage of each dataset under the unified SurgMLLMBench schema, demonstrating how heterogeneous datasets were standardized into a cohesive multimodal benchmark.

\begin{figure*}[t]
    \centering
    \includegraphics[width=1\linewidth]{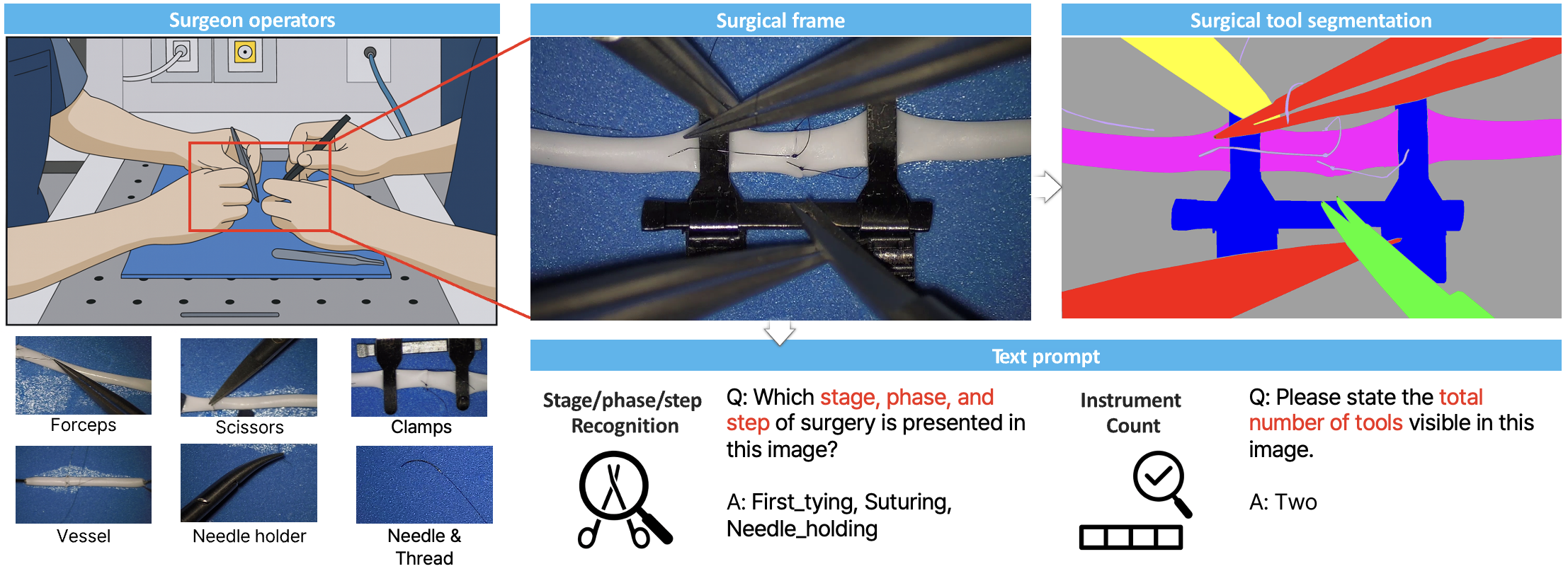}
    \caption{Overview of the MAVIS dataset collection process.}
    \label{fig:MAVIS_overview}
\end{figure*}

\vspace{0.5\baselineskip}
\noindent\textbf{MAVIS Dataset.} The MAVIS dataset is introduced as a comprehensive resource for micro-surgical workflow understanding. 
The proposed dataset targets the micro-surgery domain, where high-precision procedures demand prolonged focus from a few expert surgeons, emphasizing the role of AI and robotic assistance.
Unlike MISAW \cite{huaulme2021micro}, which includes only a single-step suturing task, our dataset captures the complete sequence of both anterior and posterior anastomosis of an artificial vessel within a single video. To enable a comprehensive understanding of the surgical workflow, we introduce a new attribute, \textit{Stage}, allowing each procedure to be hierarchically annotated following a stage–phase–step structure. The MAVIS dataset is the first to incorporate the \textit{Stage} attribute across surgical domains, providing a more detailed representation of real surgical workflows. The MAVIS dataset is available as a standalone dataset at: \href{https://huggingface.co/datasets/KIST-HARILAB/MAVIS}{huggingface.co/datasets/KIST-HARILAB/MAVIS}.

It comprises 19 videos of artificial vascular anastomosis procedures performed by three expert micro-surgeons from the College of Medicine, Korea University, who are also co-authors of this study.
The dataset includes recordings of 1 mm artificial vessel anastomosis procedures, performed by pairs of surgeons (a lead surgeon and an assistant). 
Each pair executed the full anastomosis sequence in varying orders according to their preferred surgical techniques.
Specifically, the dataset comprises seven videos from Surgeon 1, seven from Surgeon 2, and five from Surgeon 3. All videos were recorded at a resolution of 1920 × 1080 and temporally sampled at 1 FPS. For each frame, MAVIS provides pixel-level segmentation annotations for seven categories of surgical instruments and frame-level workflow annotations encompassing surgical stages, phases, and steps. The overall data collection setup and recording environment are illustrated in \cref{fig:MAVIS_overview}

\subsection{Data Annotation}
\begin{figure}[t]
    \centering
    \includegraphics[width=\linewidth]{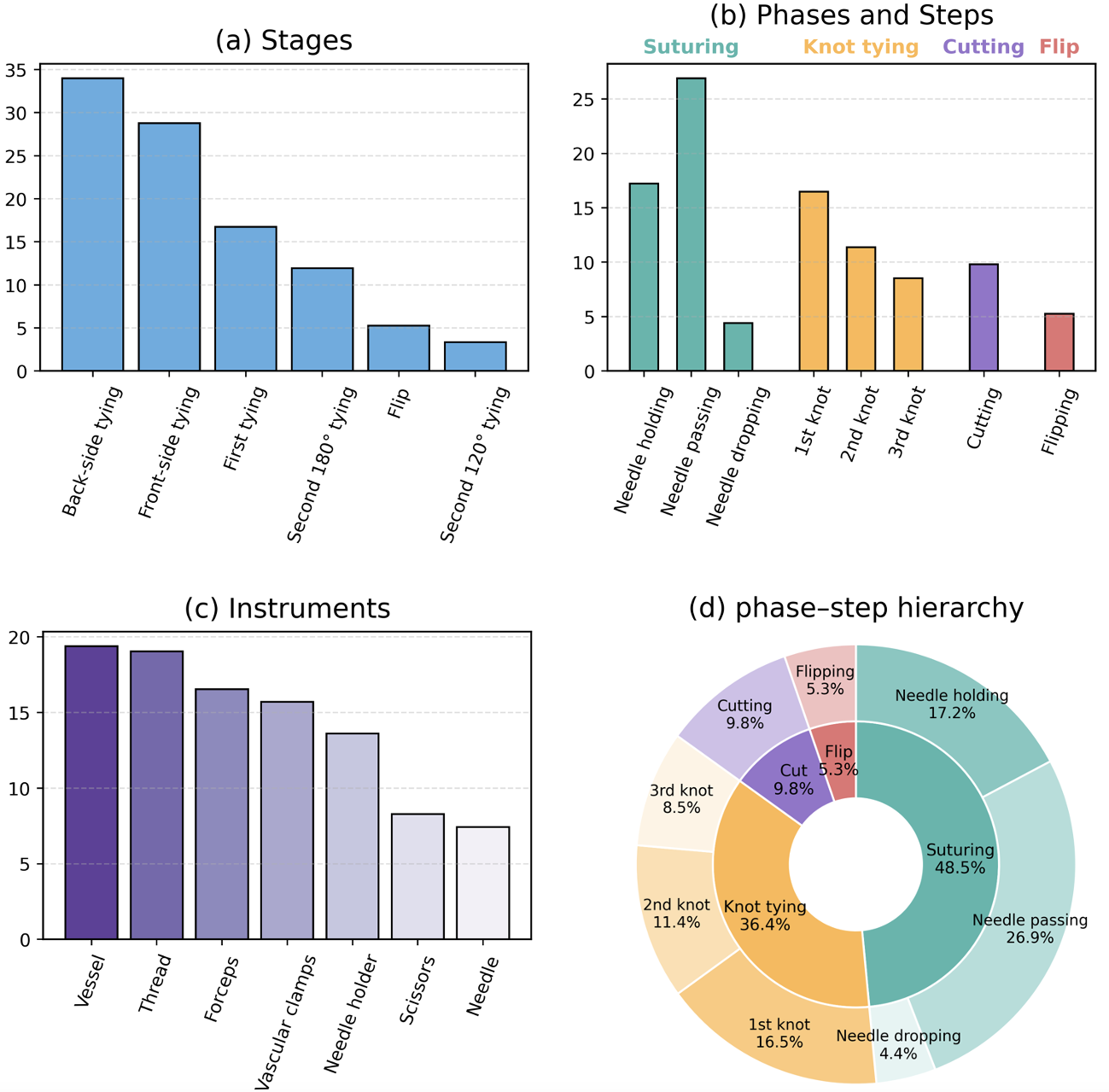}
    \caption{MAVIS dataset distributions.}
    \label{fig:mavis_dist}
\end{figure}

\noindent\textbf{MAVIS Dataset.}
The MAVIS dataset provides frame-level multi-level annotations designed for the precise analysis of micro-vascular anastomosis procedures. These annotations were constructed in collaboration with expert micro-surgeons (co-authors of this paper), reflecting real surgical workflows to ensure practical applicability in both clinical and research settings. Each frame contains workflow annotations that capture the temporal progression of surgery through a stage–phase–step hierarchy, as well as instrument segmentation annotations that represent the spatial components of the surgical scene.
The surgical workflow is organized into six major stages: \textsc{First tying}, \textsc{Second 180° tying}, \textsc{Second 120° tying}, \textsc{Front-side tying}, \textsc{Back-side tying}, and \textsc{Flip}. Each stage is subdivided into four phases—\textsc{Suturing}, \textsc{Knot tying}, \textsc{Cutting}, and \textsc{Flip}—and each phase is further decomposed into fine-grained operational steps. The step annotations include eight categories: \textsc{Needle holding}, \textsc{Needle passing}, \textsc{Needle dropping}, \textsc{Knot tying (1st–3rd knot)}, \textsc{Cutting}, and \textsc{Flipping}. This hierarchical annotation structure (see \cref{fig:mavis_dist}(d)) enables quantitative analysis of the temporal progression of surgical workflows and facilitates research on phase and step anticipation.

At the visual level, each frame includes pixel-wise segmentation annotations for all surgical instruments appearing in the scene. The annotated instrument categories comprise eight classes: \textsc{Background material}, \textsc{Forceps}, \textsc{Scissors}, \textsc{Vascular clamps}, \textsc{Needle holder}, \textsc{Vessel}, \textsc{Needle}, and \textsc{Thread}. Each object is delineated with polygonal coordinates following the COCO format, accurately capturing its spatial contour and extent. Multiple instruments may appear simultaneously within the same frame, and such pixel-level annotations can be utilized for various vision-based analyses, including instrument detection, segmentation, tracking, and action recognition \cite{minaee2020}.



\label{sec:vqa}
\vspace{0.5\baselineskip}
\noindent\textbf{VQA Prompt Generation.} To construct the VQA annotations, we combined MAVIS frame-level workflow annotations (stage, phase, step) with short-term instrument metadata and, where available from other sources (e.g., EndoVis2018, GraSP, MISAW), instrument action labels. Each frame was paired with one of several fixed prompt templates, uniformly sampled across five categories: (1) \textit{workflow queries} (e.g., “Which stage, phase, and step are shown?”), (2) \textit{instrument count queries} (e.g., “How many surgical tools are visible?”), (3) \textit{instrument type queries} (e.g., “Which instruments are present?”), (4) \textit{instrument action queries} (e.g., “What action is the needle holder performing?”), and (5) \textit{dataset source queries} (e.g., “What is the source of this dataset?”). Workflow and count prompts were always instantiated; type and action prompts were included when corresponding labels existed. The corresponding VQA prompts include dataset identifiers to help the model distinguish domain-specific contexts during training.
A template-based VQA generation approach was adopted instead of using generative language models such as GPT \cite{achiam2023gpt}. Surgical scene understanding tasks, particularly workflow recognition and instrument counting, require high precision and deterministic supervision \cite{lin2021}. Free-form question generation can introduce linguistic variability and semantic inconsistency \cite{dong2025generative}, whereas fixed prompt templates ensure consistent phrasing, reproducible dataset construction, and higher annotation accuracy, enabling stable and efficient multimodal training and evaluation \cite{awal2023investigating} within SurgMLLMBench. 

\vspace{0.5\baselineskip}
\noindent\textbf{MAVIS Dataset Statistics.} The MAVIS dataset comprises 10,652 frames, each annotated with stage, phase, and step labels. Among the six surgical stages, \textsc{Back-side tying} and \textsc{Front-side tying} occupy the largest proportions, followed by \textsc{First tying} and \textsc{Second 180° tying} (see \cref{fig:mavis_dist}(a)). At the phase and step levels, \textsc{Suturing} and \textsc{Knot tying} dominate the distribution, with \textsc{Needle passing} and \textsc{Needle holding} as the most frequent fine-grained actions (see \cref{fig:mavis_dist}(b)).
For instrument segmentation, \textsc{Vessel} and \textsc{Thread} appear most frequently, followed by \textsc{Forceps}, \textsc{Vascular clamps}, and \textsc{Needle holder}, reflecting that suturing and knot tying constitute the core operations of micro-surgical anastomosis (see \cref{fig:mavis_dist}(c)).
Overall, MAVIS presents a balanced composition of temporal workflow and visual annotations, quantitatively representing the procedural characteristics of micro-surgery. The dataset distributions are illustrated in \cref{fig:mavis_dist}.
\begin{figure}[t]
    \centering
    \begin{subfigure}{\linewidth}
        \includegraphics[height=4cm, width=1\linewidth]{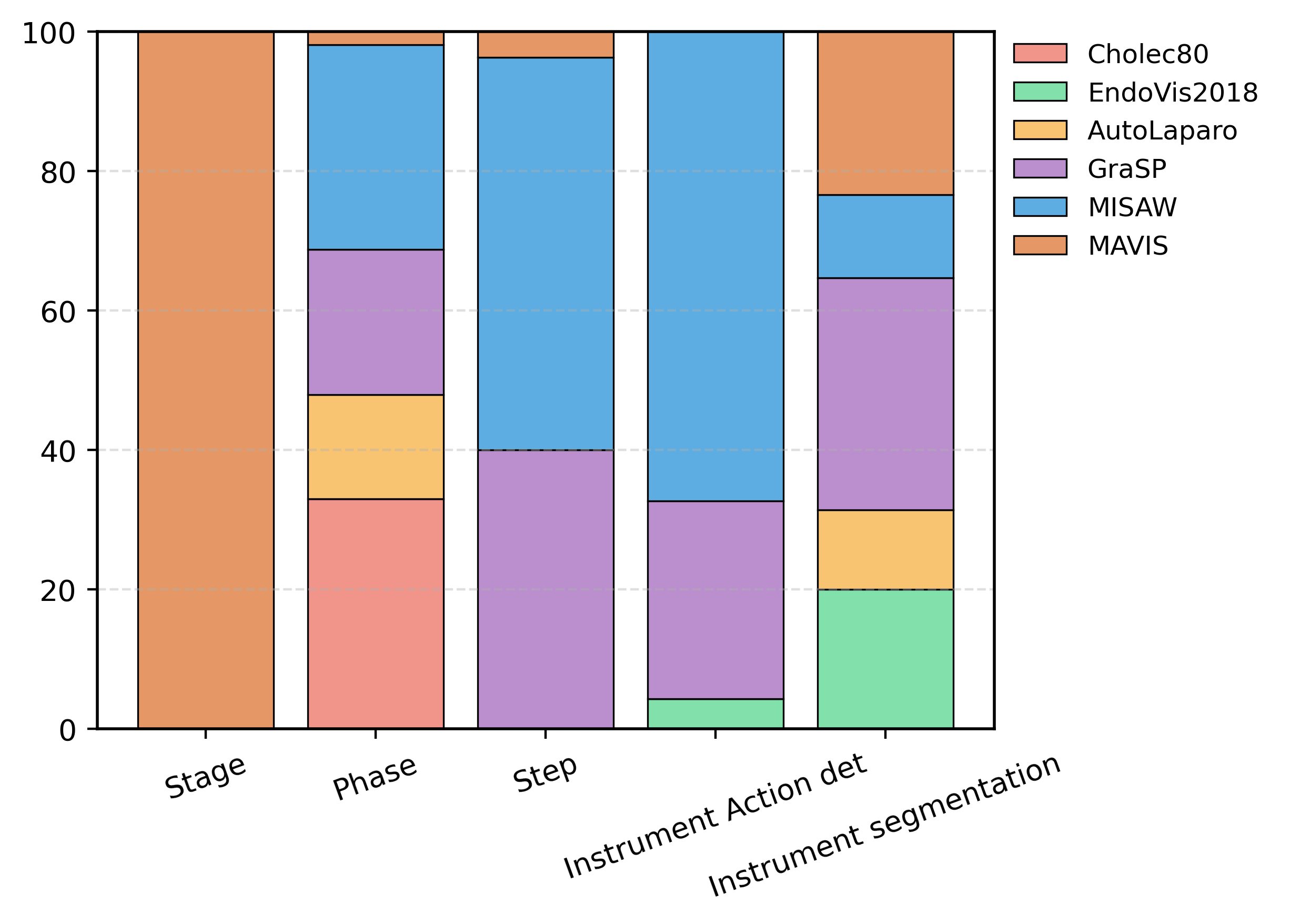}
        \caption{Dataset distribution}
    \end{subfigure}
    \begin{subfigure}{\linewidth}
        \includegraphics[height=4cm, width=0.95\linewidth]{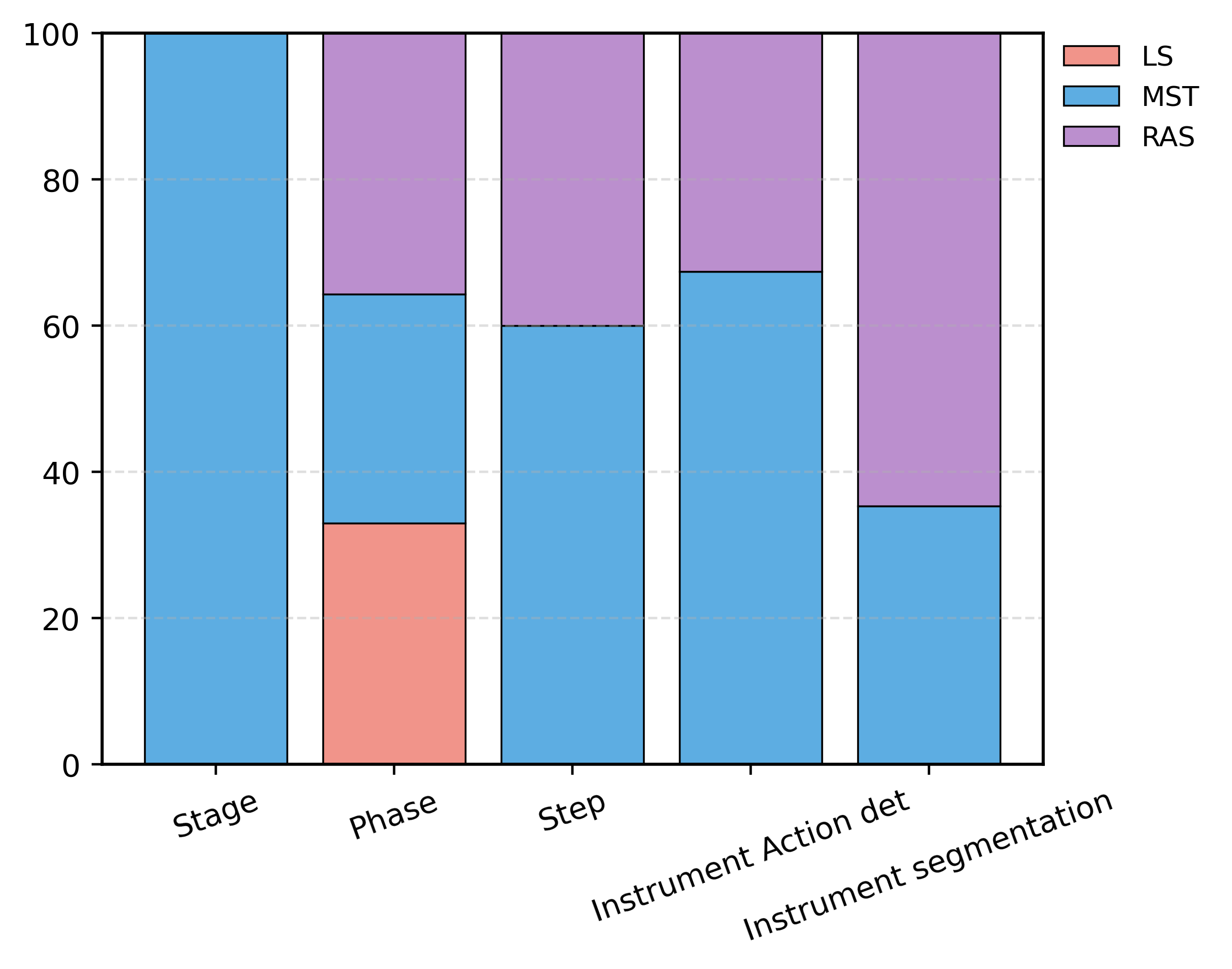}    
        \caption{Domain group distribution}
        \vspace{-0.5em}
    \end{subfigure}
    \caption{Comparison of dataset composition across tasks.}
    \label{fig:dataset_summary}
\end{figure}

\vspace{0.5\baselineskip}
\noindent\textbf{Dataset Comparison.} SurgMLLMBench integrates six surgical vision datasets: five existing datasets—Cholec80 \cite{twinanda2016endonet}, EndoVis2018 \cite{allan20202018}, AutoLaparo \cite{wang2022autolaparo}, GraSP \cite{ayobi2024pixel}, and MISAW \cite{huaulme2021micro}—plus a newly collected dataset, MAVIS. 
\cref{fig:dataset_summary} visually summarizes the composition and task coverage of each dataset within the SurgMLLMBench framework. In (a), the proportion of data dedicated to each task (stage, phase, and step recognition; instrument action detection; instrument segmentation) is shown for each dataset, while (b) illustrates the distribution of data across surgical domains (LS, MST, RAS). Stage recognition was introduced in the proposed MAVIS dataset to represent hierarchical workflow information and is only available in this dataset among the six included. For the remaining tasks, since phase recognition is only provided in Cholec80—the sole dataset in the LS domain—excluding this case, the datasets contribute roughly similar proportions of data across the different surgical domains.

\section{Performance Evaluation}
\subsection{Evaluation Methods}
\label{evaluation method}
\noindent\textbf{OMG-LLaVA.} We adopt OMG-LLaVA~\cite{zhang2024omg} as the primary interactive multimodal baseline because it can both answer open-ended surgical queries (e.g., phase, step, action, and count) and generate segmentation masks corresponding to these queries. This dual capability aligns closely with the design objectives of SurgMLLMBench, which provides workflow-level supervision alongside pixel-accurate instrument annotations across domains. Such integration enables joint learning of surgical reasoning and visual grounding within a unified multimodal framework.

Following OMG-LLaVA~\cite{zhang2024omg}, we adopt a two-stage training framework comprising pre-training for broad image–text alignment and instruction tuning for multimodal task adaptation. 
In the pre-training stage, the perception model and LLM are frozen while only the visual and text projectors are optimized to establish vision–language alignment. Since OMG-LLaVA’s pre-training stage already provides comprehensive multimodal alignment, we adopt the same pre-training dataset.
During instruction tuning, we employ LoRA~\cite{hu2022lora} to fine-tune the LLM and unfreeze the OMG-decoder to enable adaptation to surgical-specific visual features and pixel-level reasoning.
The experiments use a batch size of 4 and one training epoch, while all other hyperparameters follow the original OMG-LLaVA configuration~\cite{zhang2024omg}.

OMG-LLaVA is deliberately trained on SurgMLLMBench while excluding the MAVIS dataset. This experimental setup enables a clear assessment of cross-dataset generalization by evaluating how well the model adapts to MAVIS through additional fine-tuning, in comparison to models trained without instruction tuning. 
\vspace{0.5\baselineskip}
\noindent\textbf{LLaVA.} We include LLaVA~\cite{liu2023visual} as a text–vision baseline to quantify the effect of explicit pixel-level grounding on multimodal reasoning. Unlike OMG-LLaVA, LLaVA does not produce segmentation masks; instead, it processes image–text pairs to generate natural-language responses without a dedicated segmentation head. This makes it well-suited for recognition-oriented tasks such as phase, step, action, and count prediction.
Similar to OMG-LLaVA, we follow the two-stage training paradigm of pre-training for general image–text alignment and instruction tuning for task adaptation. 
We initialize our models from the official LLaVA checkpoint and perform instruction tuning.
Both experiments are conducted with a batch size of 16, one training epoch, and LoRA~\cite{hu2022lora} for LLM fine-tuning.
The learning rate is set to $2\times10^{-4}$ for per-dataset fine-tuning and $2\times10^{-5}$ for instruction tuning on SurgMLLMBench.
All other hyperparameters follow the original LLaVA implementation~\cite{liu2023visual}.
LLaVA is trained on SurgMLLMBench with MAVIS withheld, in order to assess the generalization ability, after which additional evaluation experiments on MAVIS are conducted.

\subsection{Evaluation Metrics}
For tasks where both ground truth and predicted outputs are textual (phase, step, action, and count), we compute accuracy as the proportion of predictions whose text exactly matches the reference label \cite{rajpurkar2016squad, everingham2010pascal}. Formally, for $N$ samples with predictions $\hat{y}_i$ and ground truths $y_i$, accuracy is defined as $\mathrm{Accuracy} = \frac{1}{N}\sum_{i=1}^{N}\mathbbm{1}[\hat{y}_i = y_i],$
where $\mathbbm{1}[\cdot]$ denotes the indicator function returning 1 when the prediction and ground truth are identical.
This strict text-matching criterion ensures that the model’s reasoning output aligns precisely with the ground truth workflow labels without relying on semantic similarity or partial credit.

For the segmentation task, we adopt the mean Intersection over Union (mIoU) metric \cite{gonzu00e1lez2020}, computed as the average IoU across all instrument classes.
For a given class $c$, IoU is defined as $\mathrm{IoU}_c = \frac{TP_c}{TP_c + FP_c + FN_c},$ where $TP_c$, $FP_c$, and $FN_c$ denote the number of true positive, false positive, and false negative pixels for class $c$, respectively.
The mIoU is then calculated as $\mathrm{mIoU} = \frac{1}{C}\sum_{c=1}^{C}\mathrm{IoU}_c,$ where $C$ is the total number of instrument classes in the dataset.
This metric quantifies the pixel-level overlap between predicted and ground truth masks and allows for direct comparison among models with segmentation capability.


\begin{table*}[t]
\centering
\small
\renewcommand{\arraystretch}{0.8}
\begin{tabular}{@{}l|c|c|c|c|c|c|c|c@{}}
\toprule
Dataset & Method & \begin{tabular}[c]{@{}c@{}}Instruction Tuning\\on SurgMLLMBench\end{tabular} & \begin{tabular}[c]{@{}c@{}}Fine-tuning\\on each dataset\end{tabular} & Phase & Step & Action & Count & \begin{tabular}[c]{@{}c@{}}Instrument\\ Segmentation\end{tabular} \\ \midrule
\multirow{4}{*}{Cholec80} & LLaVA & \xmark & \cmark & 81.98 & \cellcolor{gray!50} & \cellcolor{gray!50} & 81.54 & \cellcolor{gray!50} \\
                          & LLaVA\textsuperscript{\textsection} & \cmark & \xmark & 77.32 & \cellcolor{gray!50} & \cellcolor{gray!50} & 78.73 & \cellcolor{gray!50} \\
                          & OMG-LLaVA & \xmark & \cmark & 76.51 & \cellcolor{gray!50} & \cellcolor{gray!50} & 83.08 & \cellcolor{gray!50} \\
                          & OMG-LLaVA\textsuperscript{\textsection} & \cmark & \xmark & 55.78 & \cellcolor{gray!50} & \cellcolor{gray!50} & 82.53 & \cellcolor{gray!50} \\ \midrule
\multirow{4}{*}{EndoVis2018} & LLaVA & \xmark & \cmark &                                 \cellcolor{gray!50} & \cellcolor{gray!50}                                & 34.31 & 58.88 & - \\
                              & LLaVA\textsuperscript{\textsection} & \cmark & \xmark &  \cellcolor{gray!50} & \cellcolor{gray!50}  & 43.65 & 78.38 & - \\
                              & OMG-LLaVA & \xmark & \cmark & \cellcolor{gray!50} & \cellcolor{gray!50} & 45.47 & 60.63 & 26.04 \\
                              & OMG-LLaVA\textsuperscript{\textsection} & \cmark & \xmark & \cellcolor{gray!50} & \cellcolor{gray!50} & 44.53 & 63.09 & 27.23 \\ \midrule
\multirow{4}{*}{AutoLaparo}   & LLaVA & \xmark & \cmark & 23.00 & \cellcolor{gray!50} & \cellcolor{gray!50} & 57.21 & - \\
                              & LLaVA\textsuperscript{\textsection} & \cmark & \xmark & 24.67 & \cellcolor{gray!50} & \cellcolor{gray!50} & 82.56 & - \\
                              & OMG-LLaVA & \xmark & \cmark & 22.64 & \cellcolor{gray!50} & \cellcolor{gray!50} & 67.91 & 59.57 \\
                              & OMG-LLaVA\textsuperscript{\textsection} & \cmark & \xmark & 26.79 & \cellcolor{gray!50} & \cellcolor{gray!50} & 57.91 & 44.97 \\ \midrule
\multirow{4}{*}{MISAW}        & LLaVA & \xmark & \cmark & 87.16 & 64.03 & 48.58 & 94.17 & - \\
                              & LLaVA\textsuperscript{\textsection} & \cmark & \xmark & 88.20 & 58.69 & 19.73 & 94.70 & - \\
                              & OMG-LLaVA & \xmark & \cmark &86.44 & 62.85 & 58.49 & 93.46 & 61.51 \\
                              & OMG-LLaVA\textsuperscript{\textsection} & \cmark & \xmark & 86.33 & 56.56 & 48.68 & 83.57 & 59.06 \\ \midrule
\multirow{4}{*}{GraSP}        & LLaVA & \xmark & \cmark &  68.83 & 52.75 & 39.95 & 54.40 & - \\
                              & LLaVA\textsuperscript{\textsection} & \cmark & \xmark & 66.78 & 50.83 & 2.95 & 47.56 & - \\
                              & OMG-LLaVA & \xmark & \cmark & 47.85 & 37.26 & 33.37 & 55.02 & 66.65 \\
                              & OMG-LLaVA\textsuperscript{\textsection} & \cmark & \xmark & 36.55 & 31.84 & 43.69 & 44.80 & 53.06 \\ \bottomrule

\end{tabular}
\caption{Comparison of multimodal LLM performance across surgical tasks and datasets. Gray cells indicate the absence of corresponding annotations in each dataset. \textsuperscript{\textsection}Single instruction-tuned model per method (no additional fine-tuning).}
\label{main_result}
\end{table*}

\begin{table*}[t]
\centering
\small
\renewcommand{\arraystretch}{0.8}
\begin{tabular}{@{}c|c|c|c|c|c|c|c@{}}
\toprule
 Method & \begin{tabular}[c]{@{}c@{}}Instruction Tuning\\on SurgMLLMBench\end{tabular} & \begin{tabular}[c]{@{}c@{}}Fine-tuning\\on MAVIS\end{tabular} & Stage & Phase & Step & Count & \begin{tabular}[c]{@{}c@{}}Instrument\\ Segmentation\end{tabular} \\ \midrule
\multirow{2}{*}{LLaVA} & \xmark & \cmark & 67.67 & 65.09 & 37.59 & 54.67 & - \\
 & \cmark & \cmark & 75.70 & 69.43 & 44.80 & 68.47 & - \\
 \midrule
\multirow{2}{*}{OMG-LLaVA} & \xmark & \cmark & 62.84 & 63.63 & 37.28 & 55.46 & 53.96 \\
 & \cmark & \cmark & 43.90 & 55.72 & 29.48 & 47.51 & 36.89 \\
 \bottomrule
\end{tabular}
\caption{Results of multimodal LLMs on the MAVIS dataset, which was excluded from SurgMLLMBench instruction tuning to assess model adaptability to unseen surgical dataset. 
}
\label{mavis_result}
\end{table*}

\subsection{Evaluation Results}

\noindent\textbf{SurgMLLMBench Instruction Tuning Performance.} \cref{main_result} summarizes the performance of multimodal LLMs (LLaVA~\cite{liu2023visual} and OMG-LLaVA~\cite{zhang2024omg}) on five datasets within the SurgMLLMBench. 
Each model is evaluated under two training configurations: (1) \textit{fine-tuning on each dataset}, where the model is trained independently on each dataset, and (2) \textit{instruction tuning on SurgMLLMBench}, where the model is trained once on the full SurgMLLMBench corpus excluding the MAVIS dataset and evaluated directly on all datasets without additional fine-tuning. This comparison highlights the generalization and scalability of a unified multimodal LLM trained across multiple surgical domains.

\cref{main_result} demonstrates that a single model instruction-tuned on the full SurgMLLMBench corpus achieves competitive performance across diverse surgical datasets, despite not being optimized for any specific domain. In particular, for EndoVis2018 and AutoLaparo, the model trained on SurgMLLMBench performs comparably to, or even surpasses, models fine-tuned individually on each dataset. Although some dataset-specific models exhibit slightly higher scores in their native domains, this is likely attributable to overfitting to dataset-specific label semantics or visual styles, and the performance differences remain modest. 
These results suggest that performing large-scale, cross-domain instruction tuning on SurgMLLMBench enables consistently strong performance across heterogeneous surgical environments, without incurring significant accuracy degradation. 
Notably, the findings highlight the effectiveness of SurgMLLMBench in facilitating the development of a single model that can perform interactive VQA across multiple surgical domains.

\begin{figure*}[t]
\centering
\begin{subfigure}[t]{0.49\textwidth} 
\centering
\resizebox{\textwidth}{!}{
  \renewcommand{\arraystretch}{0.2}
  \setlength{\tabcolsep}{1pt}
  \begin{tabular}{c|ccc}
    \includegraphics[width=0.2\textwidth]{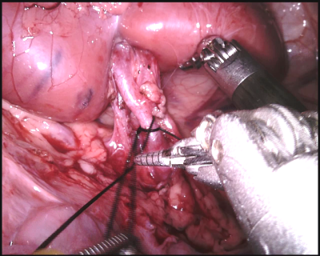}  &
    \includegraphics[width=0.2\textwidth]{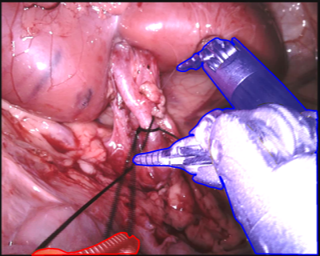}   &
    \includegraphics[width=0.2\textwidth]{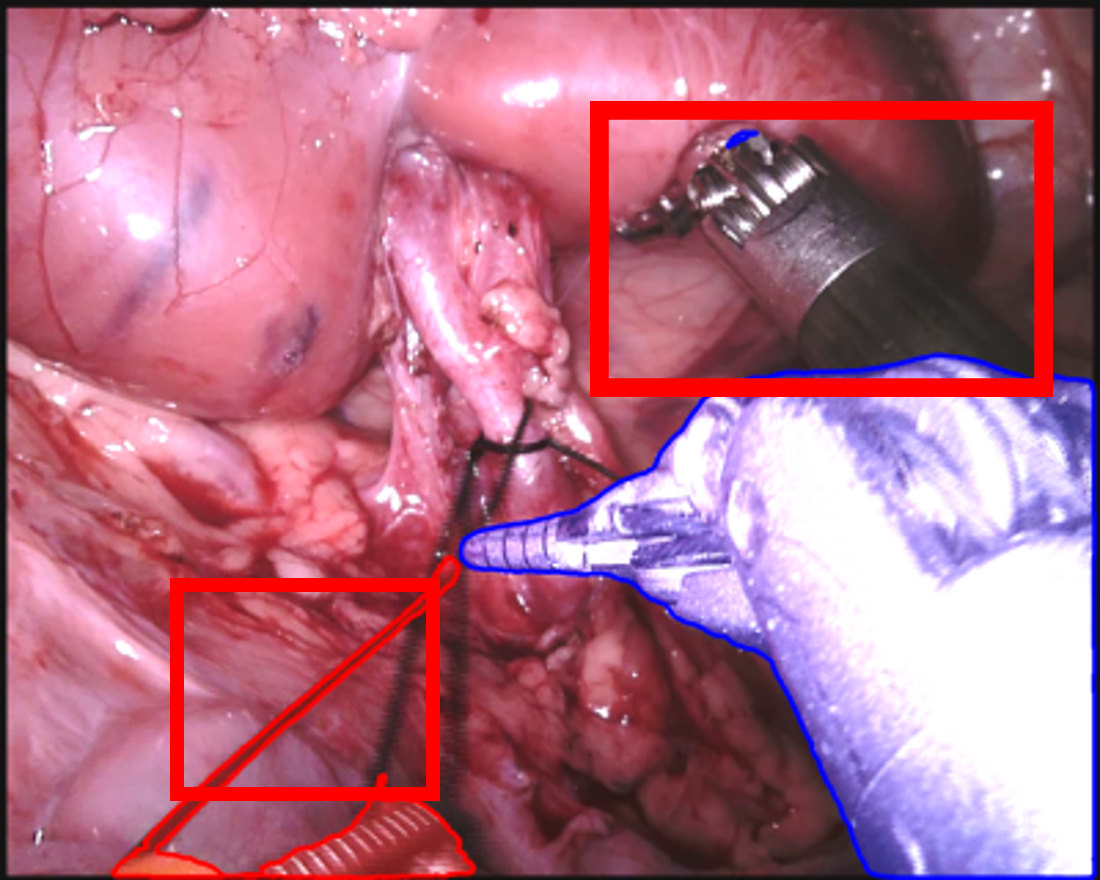} & 
    \includegraphics[width=0.2\textwidth]{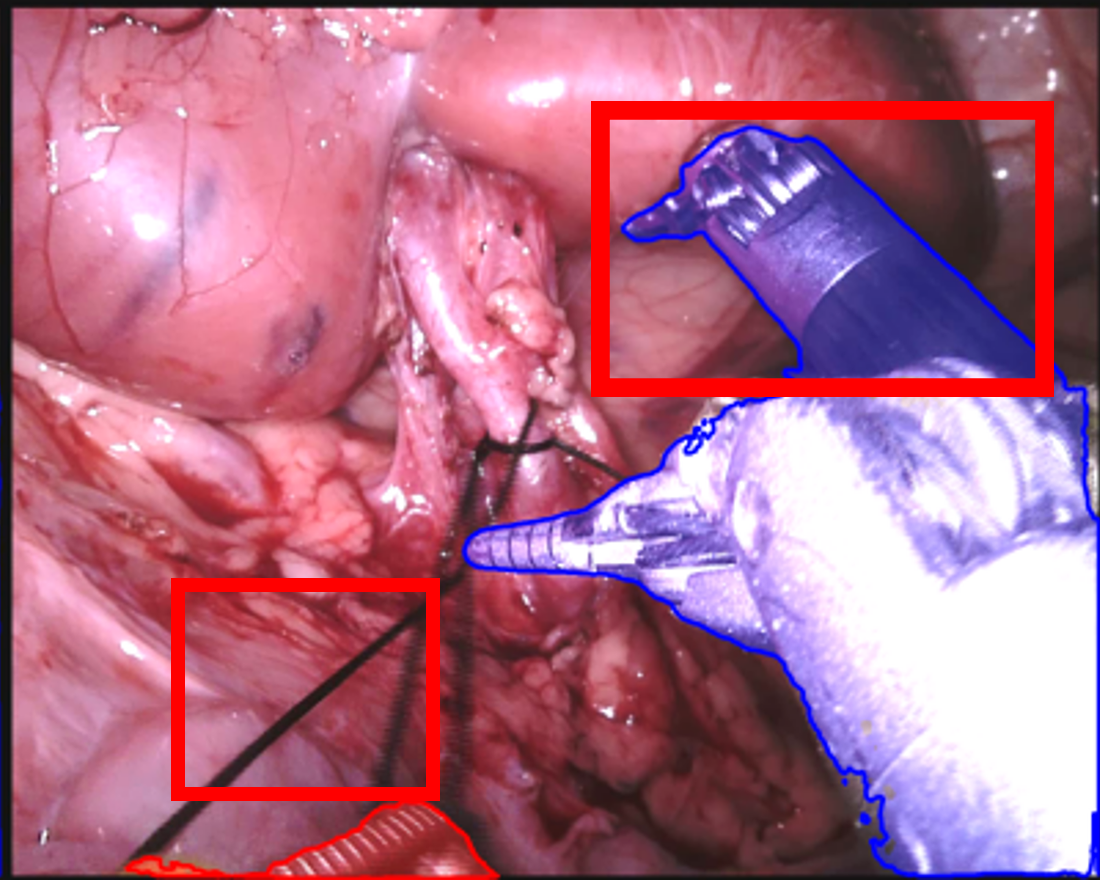}  \\

    \includegraphics[width=0.2\textwidth]{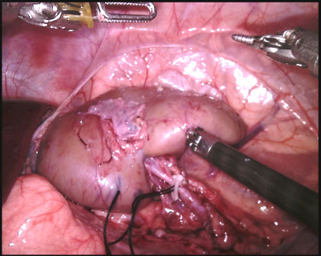}  &
    \includegraphics[width=0.2\textwidth]{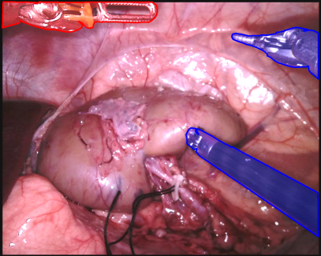}   &
    \includegraphics[width=0.2\textwidth]{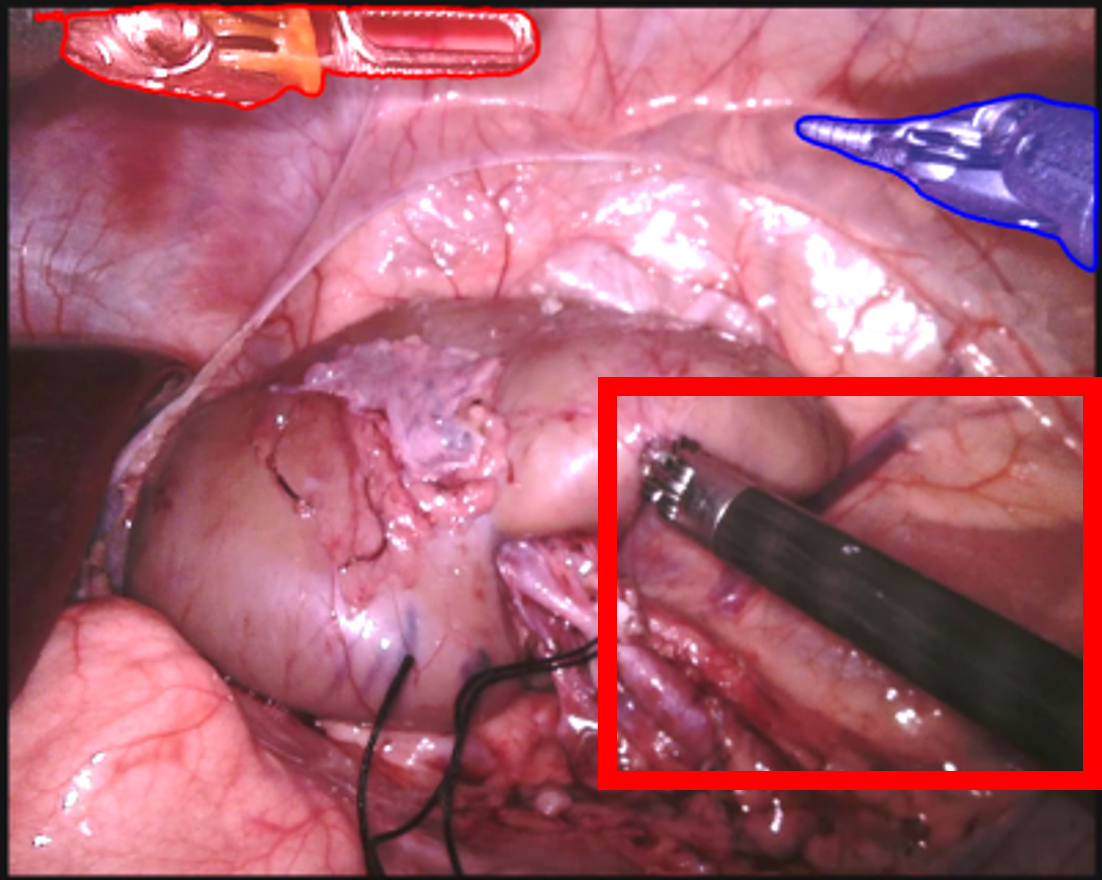} & 
    \includegraphics[width=0.2\textwidth]{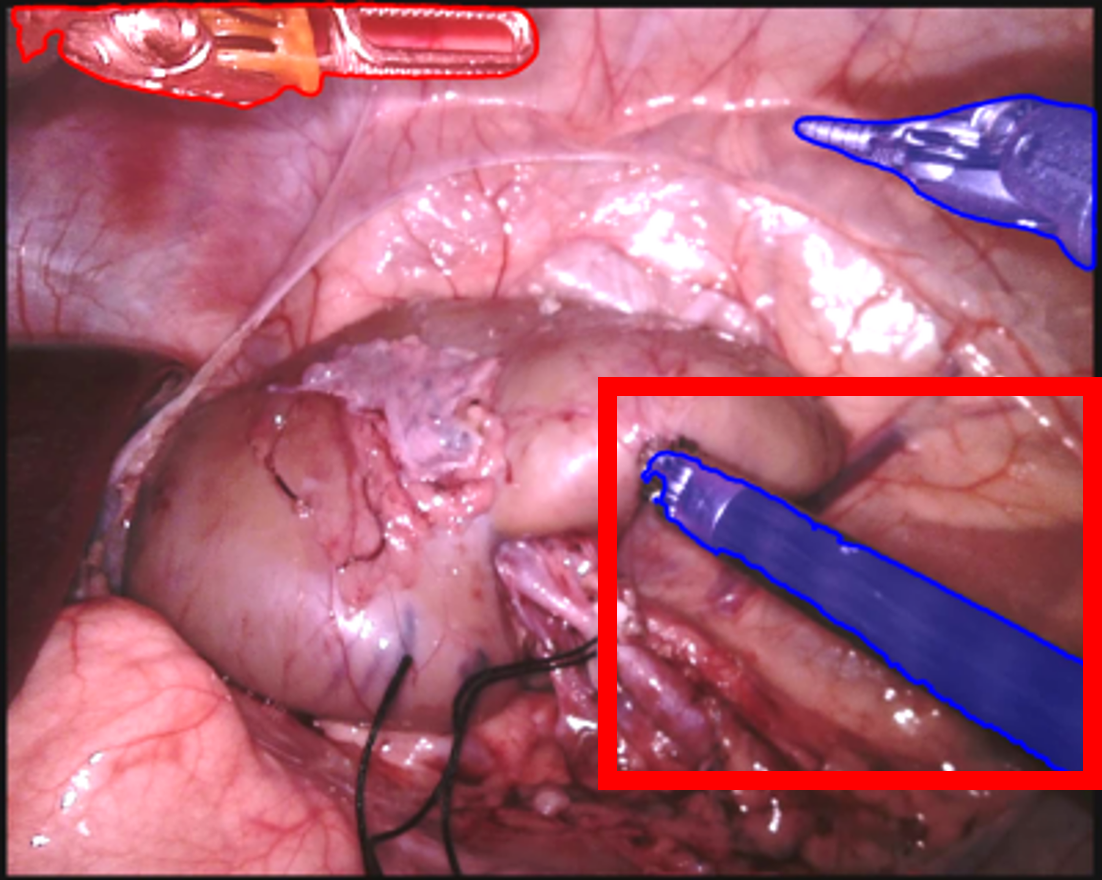}  \\

    \includegraphics[width=0.2\textwidth]{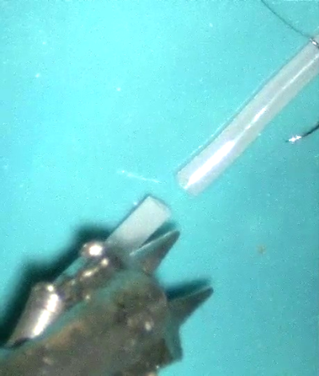}  &
    \includegraphics[width=0.2\textwidth]{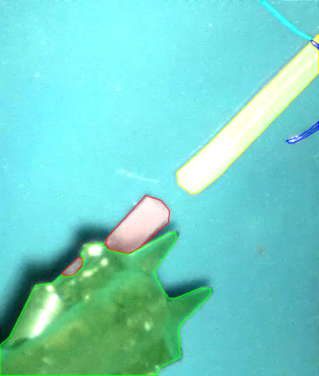}   &
    \includegraphics[width=0.2\textwidth]{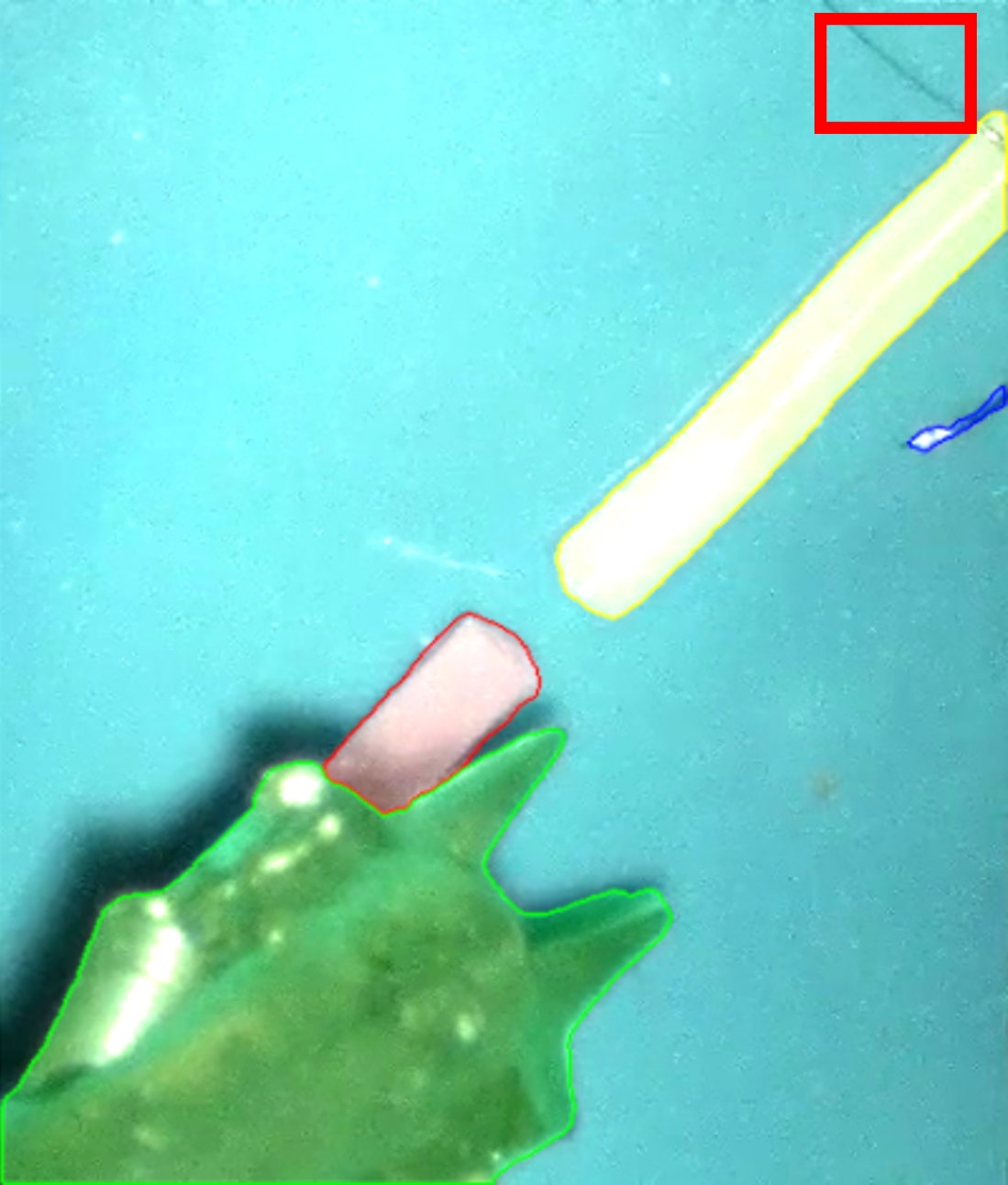} & 
    \includegraphics[width=0.2\textwidth]{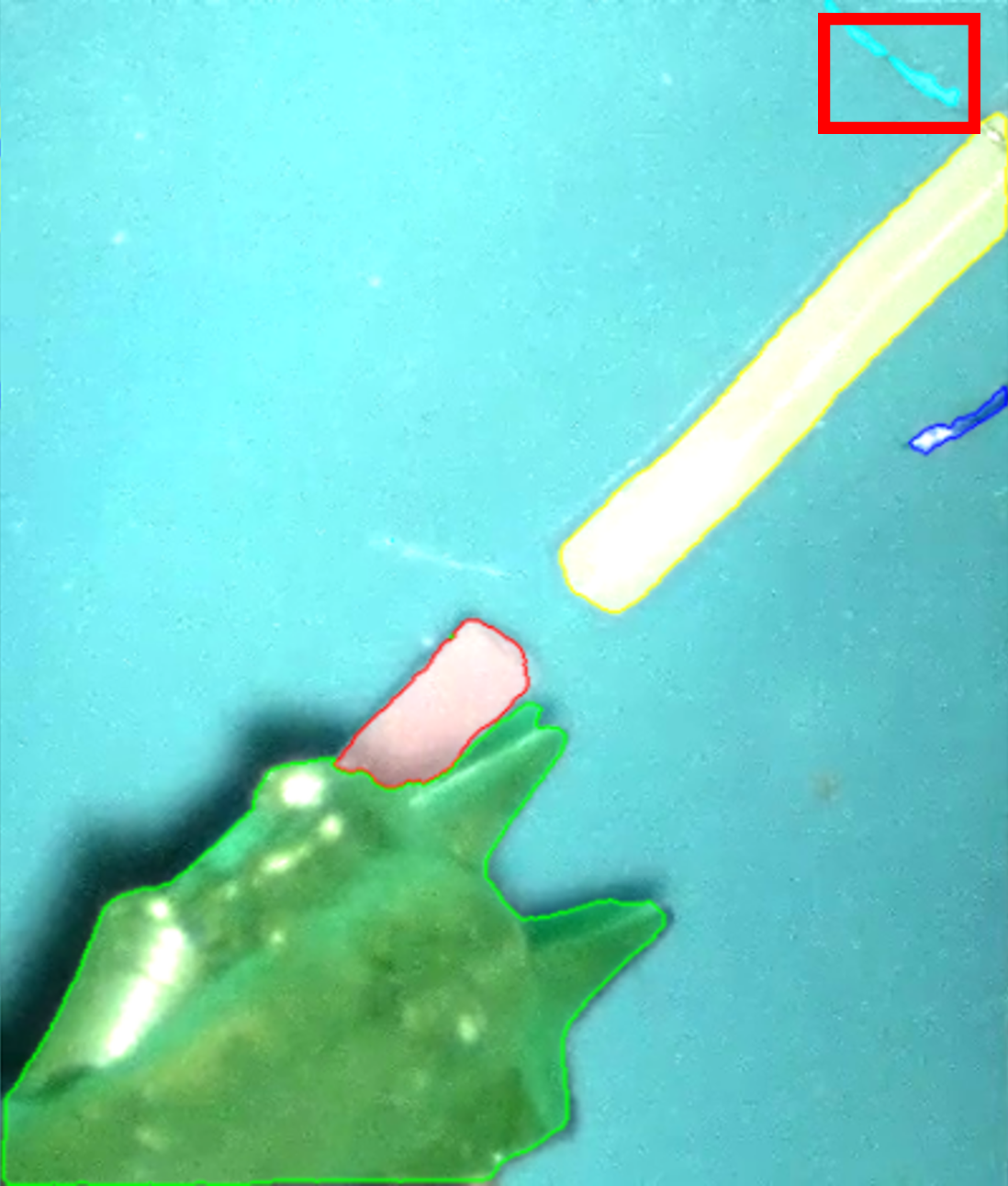}  \\

    \includegraphics[width=0.2\textwidth]{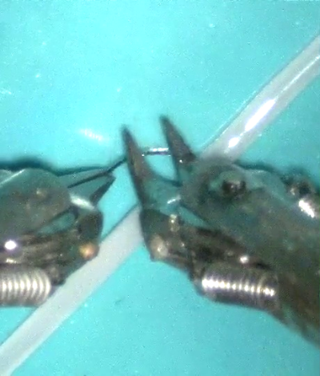}  &
    \includegraphics[width=0.2\textwidth]{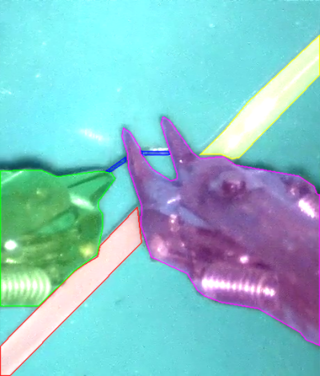}   &
    \includegraphics[width=0.2\textwidth]{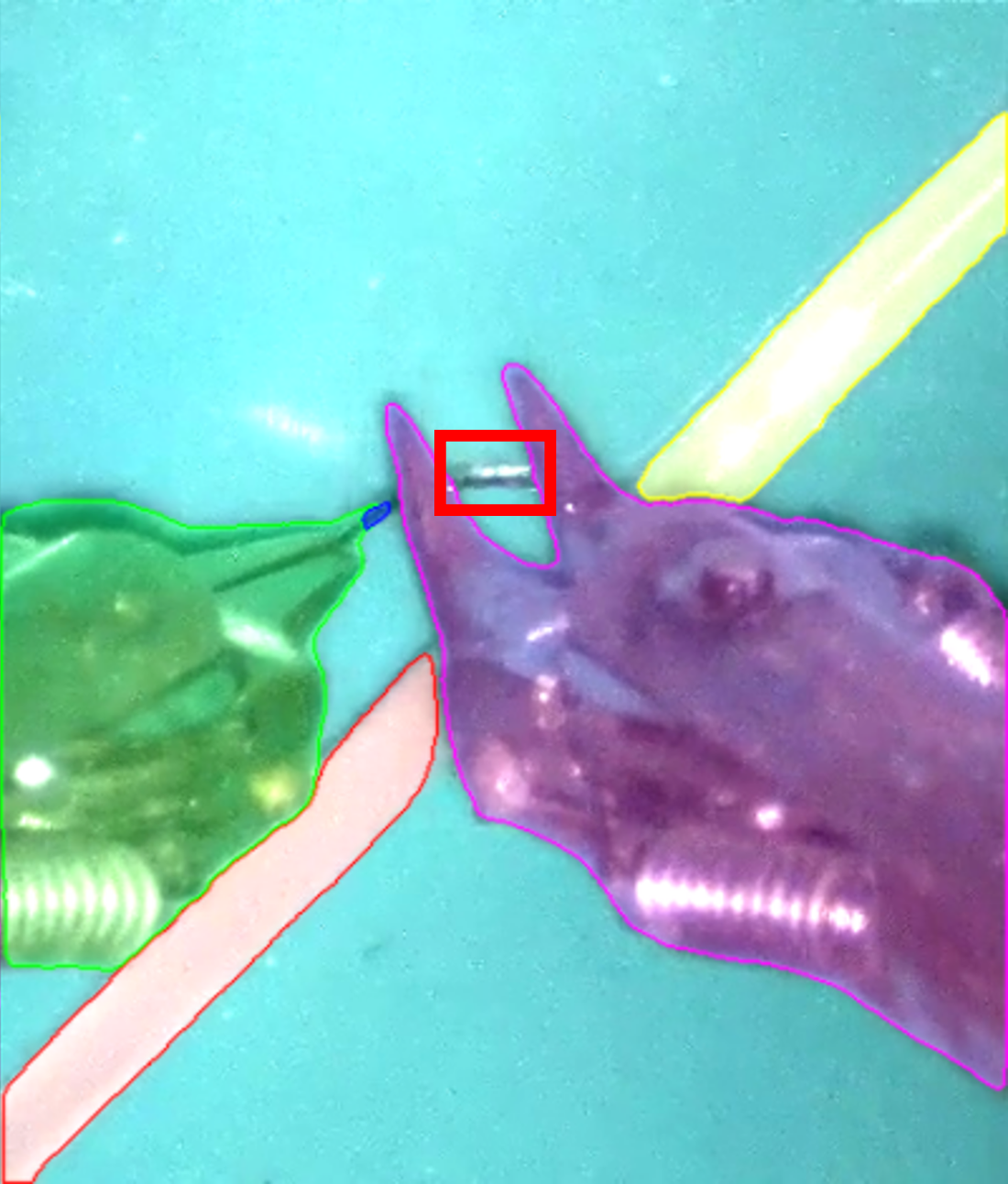} & 
    \includegraphics[width=0.2\textwidth]{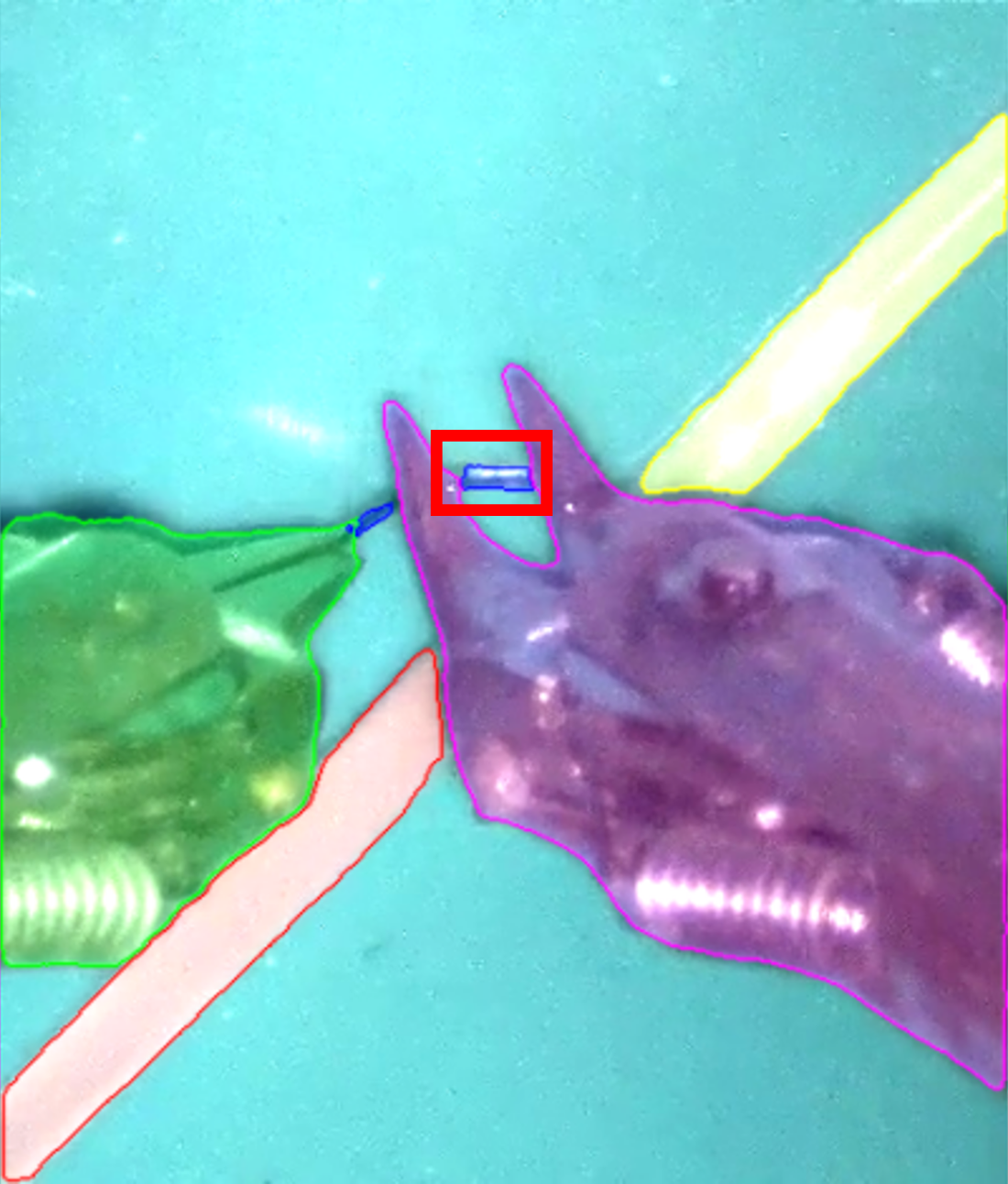}  \\
  
  \scriptsize Original Image & \scriptsize GT & \scriptsize OMG-LLaVA & \scriptsize OMG-LLaVA\textsuperscript{\textsection} 
  \end{tabular}}

\renewcommand{\arraystretch}{1}
\subcaption{Instrument segmentation comparison}
\label{fig:omg_grid}
\end{subfigure}
\hfill
\begin{subfigure}[t]{0.5\textwidth} 
\centering
\includegraphics[width=\textwidth]{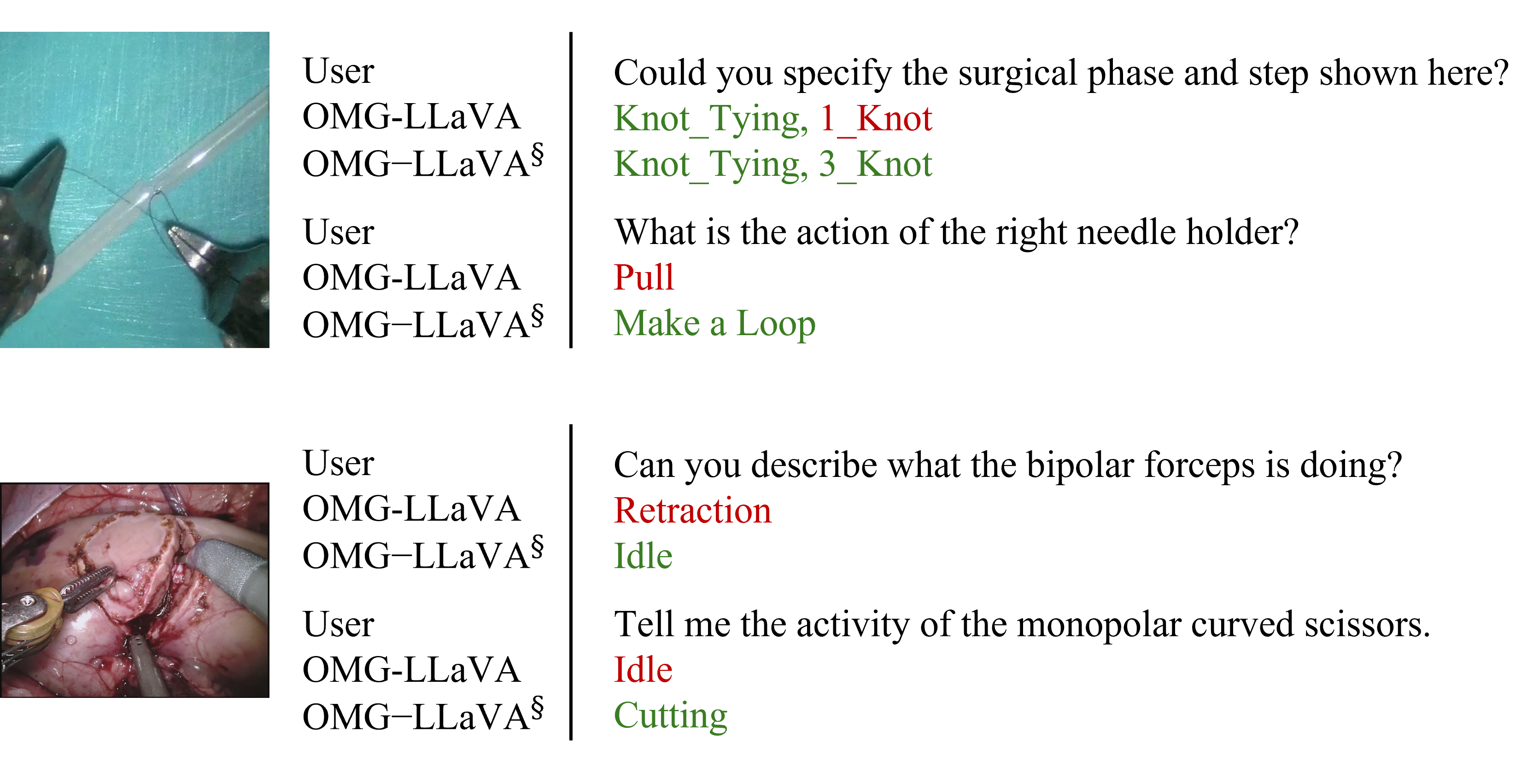}
\subcaption{Interactive workflow recognition by OMG-LLaVA}
\vspace{2mm}
\includegraphics[width=\textwidth]{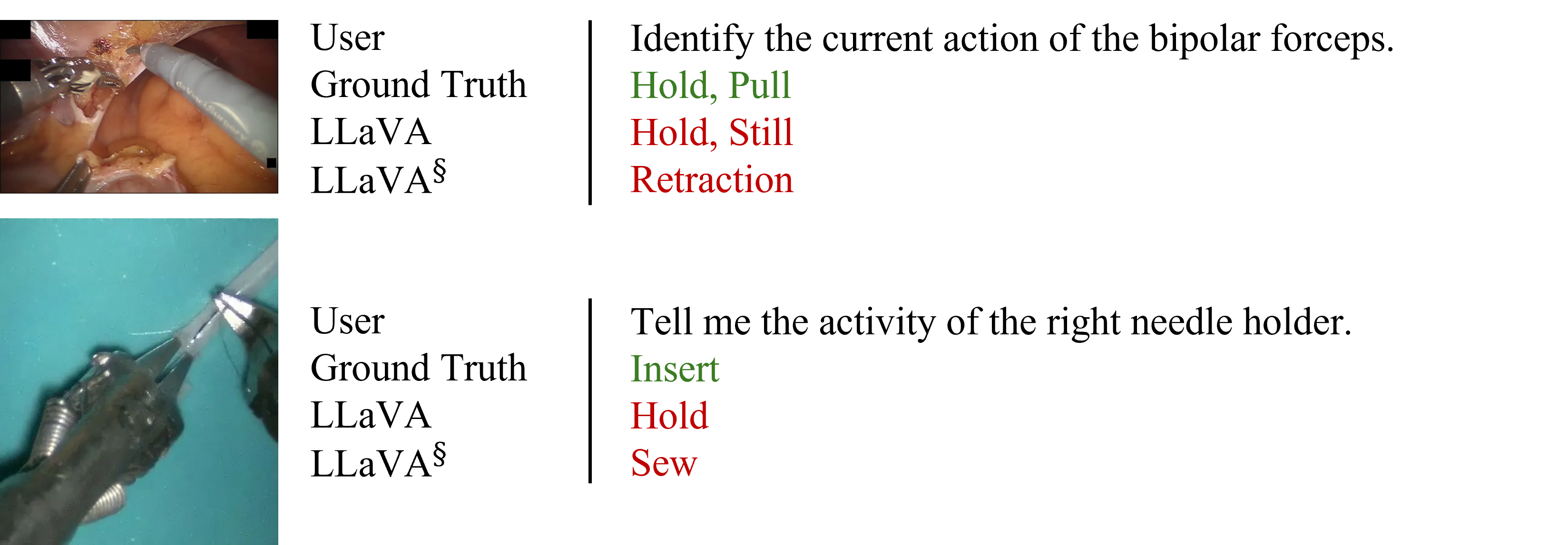}
\subcaption{\centering Interactive action detection by LLaVA \\(LLaVA\textsuperscript{\textsection} provides an incorrect but contextually valid answer)}
\end{subfigure}

\vspace{-1mm}
\caption{Qualitative visualization results of (a) instrument segmentation and (b, c) workflow recognition via VQA (green: correct, red: incorrect). OMG-LLaVA and LLaVA denote models trained individually on each dataset, whereas OMG-LLaVA\textsuperscript{\textsection} and LLaVA\textsuperscript{\textsection} represent a single model trained on SurgMLLMBench without additional per-dataset fine-tuning.
}
\label{visualization}
\end{figure*}

For GraSP and MISAW, LLaVa trained on SurgMLLMBench exhibits lower action prediction accuracy. This degradation stems from the imbalanced task distribution within the benchmark—although the number of action labels is large, the actual sample count per label is limited. As a result, the model tends to conflate semantically related actions across datasets (e.g., \textsc{Idle} vs. \textsc{Still}, \textsc{Retraction} vs. \textsc{Pull}), where functionally similar gestures are expressed under different naming conventions (see \cref{visualization}(c)). We argue that this imbalance leads to reduced numerical accuracy, even when the predicted actions remain contextually valid.

\vspace{0.5\baselineskip}
\noindent\textbf{Generalization Ability.} To further assess the generalization ability of instruction tuning on the entire SurgMLLMBench, we evaluate all models on the MAVIS dataset, which is intentionally excluded from the instruction tuning corpus (\cref{mavis_result}). This setup allows us to test whether models trained on SurgMLLMBench can adapt to unseen datasets and novel tasks, such as stage recognition. In this evaluation, the instruction-tuned model\textsuperscript{\textsection} from \cref{main_result} serves as the initialization for additional fine-tuning on MAVIS and is compared against models trained solely on the MAVIS dataset.  
The checkpoint trained on SurgMLLMBench is further fine-tuned on MAVIS (3 epochs for LLaVA and 1 epoch for OMG-LLaVA), enabling evaluation of feature transfer to unseen datasets.

LLaVA benefits from SurgMLLMBench initialization, achieving consistent improvements across tasks and demonstrating strong adaptability even when transferred to a new dataset that introduces a novel task, stage recognition.
These results indicate that performing large-scale cross-domain instruction tuning on SurgMLLMBench enhances the ability of text–vision reasoning models to generalize to unseen tasks and datasets.
For OMG-LLaVA, we observe that initializing from the model trained on SurgMLLMBench can result in performance degradation, likely due to visual gaps between datasets limiting the transferability of its pixel-level decoder. 
\vspace{0.5\baselineskip}
\noindent\textbf{Visualization.} \cref{visualization}(a) presents qualitative results of OMG-LLaVA predictions under two training strategies: dataset-specific fine-tuning and SurgMLLMBench instruction tuning. The top two rows show results from the EndoVis2018 dataset~\cite{allan20202018}, while the bottom two rows correspond to the MISAW dataset~\cite{huaulme2021micro}.
In the EndoVis2018 results, the model fine-tuned only on EndoVis2018 often misses existing instrument masks or produces spurious background predictions, suggesting overfitting to dataset-specific visual patterns. Conversely, the model trained on SurgMLLMBench—exposed to more diverse surgical scenes—shows more stable and conservative segmentation, reducing false detections and improving mask consistency.
In MISAW, the SurgMLLMBench-trained model more accurately captures thin and small structures, such as needles and wires, compared to the MISAW-only fine-tuned model. This indicates that cross-domain instruction tuning encourages more generalized visual representations, improving recognition of delicate, low-contrast surgical tools.

\cref{visualization}(b) shows that OMG-LLaVA instruction-tuned on SurgMLLMBench performs more accurate workflow recognition on MISAW and EndoVis2018 through VQA than the dataset-specific model, evidencing stronger workflow-aware visual reasoning. Conversely, \cref{visualization}(c) illustrates the LLaVA’s action recognition result on MISAW and GraSP: the SurgMLLMBench-tuned model predicts the label from the Endovis2018 dataset (\textsc{Retraction}) instead of \textsc{Pull}, revealing cross-dataset semantic alignment that is contextually valid yet penalized by dataset-specific label names.
Overall, these visual comparisons demonstrate that instruction tuning on the entire SurgMLLMBench improves both visual grounding and workflow reasoning, and achieves more consistent VQA predictions across tasks while maintaining semantic alignment even under cross-dataset label variations.

\section{Conclusion}
We introduced SurgMLLMBench, a unified benchmark for interactive multimodal LLMs that integrates workflow annotations (stage, phase, and step) with pixel-level instrument masks across multiple surgical domains, including the newly proposed MAVIS dataset. Baseline experiments demonstrate that training on SurgMLLMBench yields stable multimodal reasoning, enables pixel-accurate visual grounding, and improves robustness under domain shifts.
Beyond serving as a benchmark, SurgMLLMBench provides a practical foundation for training multimodal surgical assistants that can answer textual queries and provide visual evidence via segmentation masks. It also enables cross-domain generalization analysis and supports the development of agentic systems that maintain conversational context, leverage vision tools, and offer interpretable visual explanations—advancing applications in surgical education, intraoperative assistance, and robotic autonomy. 

Future work will extend temporal and multimodal coverage (e.g., kinematics, audio, depth), broaden action taxonomies and instance tracking, introduce topology-aware metrics, and assess real-time reliability, uncertainty, and safety for clinical deployment.

{
    \small    
    \bibliographystyle{ieeenat_fullname}
    \bibliography{main}
}


\end{document}